\documentclass{article}
\usepackage{rsj2018e_IS}
\usepackage{enumitem}
\usepackage{amsmath,graphicx}
\usepackage{bm}
\usepackage{amssymb}
\usepackage{amsfonts}

\usepackage{hyperref}
\usepackage{color}
\usepackage{times}
\usepackage{epsfig}
\usepackage{epstopdf}
\usepackage{caption}
\usepackage{graphicx}
\usepackage{amsmath}
\usepackage{amssymb}
\usepackage{subfigure}
\usepackage{latexsym}

\usepackage{amsmath}

\usepackage{siunitx}

\usepackage[linesnumbered, ruled, vlined]{algorithm2e}  
\usepackage[capitalise]{cleveref}                       

\usepackage{todo}

\usepackage{algorithmic}

\usepackage{array}
\newcolumntype{L}[1]{>{\raggedright\let\newline\\\arraybackslash\hspace{0pt}}m{#1}}
\newcolumntype{C}[1]{>{\centering\let\newline\\\arraybackslash\hspace{0pt}}m{#1}}
\newcolumntype{R}[1]{>{\raggedleft\let\newline\\\arraybackslash\hspace{0pt}}m{#1}}

\begin{document}

\title{
Experimental Force-Torque Dataset for Robot Learning of Multi-Shape Insertion
}

\author{Giovanni De Magistris$^{1}$, Asim Munawar$^{1}$, Tu-Hoa Pham$^{1}$, Tadanobu Inoue$^{1}$,
\\
Phongtharin Vinayavekhin$^{1}$, Ryuki Tachibana$^{1}$
\\
$^{1}$IBM Research - Tokyo, Japan} 

\engtitle{}
\engauthor{}

\setlength{\baselineskip}{4.4mm}
\maketitle
\thispagestyle{empty}
\pagestyle{empty}

\maketitle


\textit{
    The accurate modeling of real-world systems and physical interactions is a common
    challenge towards the resolution of robotics tasks.
    Machine learning approaches have demonstrated significant results
    in the modeling of complex systems
    (e.g., articulated robot structures, cable stretch, fluid dynamics),
    or to learn robotics tasks (e.g., grasping, reaching)
    from raw sensor measurements without explicit programming,
    using reinforcement learning.
    However, a common bottleneck in machine learning techniques
    resides in the availability of suitable data.
    While many vision-based datasets have been released in the recent years,
    ones 
    involving physical interactions,
    of particular interest for the robotic community,
    have been scarcer.
    In this paper,
    we present a public dataset on peg-in-hole insertion tasks
    containing force-torque and pose information for multiple variations of
    convex-shaped pegs. 
    We demonstrate how this dataset can be used
    to train a robot to insert polyhedral
    pegs into holes using only 6-axis force/torque sensor measurements as inputs,
    as well as other tasks involving contact such as shape recognition.
}

\section{Introduction}
\label{sec:intro}
Robot manufacturers are focused on making robots simpler to program to speed up the configuration of new assembly lines. 
Owing to recent advances in deep learning and machine learning, robots are becoming more flexible.
Instead of manual programming, modern artificial intelligence allows robots to learn new tasks by looking at demonstrations or actively learning without explicit teaching.
Recent works have already shown the potential to learn the robot dynamics~\cite{icra:pham:2018} or learn contact dynamics during a peg-in-the-hole task~\cite{conf:iros:inoue2017}.

Data is a key to the success of machine learning for solving complex tasks. 
The emergence of large datasets has played a prominent role in different research communities where deep learning has provided state-of-the-art results, e.g. natural language processing~\cite{journals:corr:HewlettLJPFHKB16}, image and score understanding~\cite{conf:cvpr:Cordts2016Cityscapes,Shahroudy_2016_CVPR}.
%
%
The robotic community still lacks public datasets, especially for problems that are complex to model like contact tasks, where it is still difficult to obtain a precise model of the physical interaction between two objects~\cite{conf:gi:bouchard2015}.
Therefore, we believe that availability of more datasets collected using real robots is crucial. 
Towards this ambitious goal, Yu \textit{et al.}~\cite{conf:iros:yu2016} is one of the first works to provide a large dataset on a robot contact task, with force information during pushing task.

In this paper, we choose one of the most common industrial tasks: the peg-in-hole task. 
We provide a dataset of a force/torque (F/T) data of peg-in-hole operations with polyhedral pegs and holes. 
If the robot has a precise position control and the hole pose is estimated with enough accuracy, we can solve this problem using position commands. However, usually due to uncertainty of robotic assembly, the task becomes unsolvable by positioning alone; the sources of the uncertainty include object positioning errors, hole pose estimation inaccuracy and grasping inaccuracy. Hence, in this paper, we put emphasis on the F/T data of the task. 
The F/T dataset presented in this paper allow to assess the feasibility of novel techniques before further effort to realize them physically or to help pre-train neural networks for insertion tasks or shape recognition.

\section{Force-Based Insertion Dataset}
\label{sec:dataset}
In this paper, we choose a strategy to solve the peg-in-hole~\cite{conf:iros:inoue2017}: i) position the peg at a predefined height from the hole, ii) push the peg with a downwards force, iii) place the peg center within the clearance region of the hole center applying force/torque movements (search phase) and iv) push the peg with a downwards force (insertion phase).

\subsection{Data collection} 
\begin{figure}[htpb]
\centering
\includegraphics[width=1.0\columnwidth]{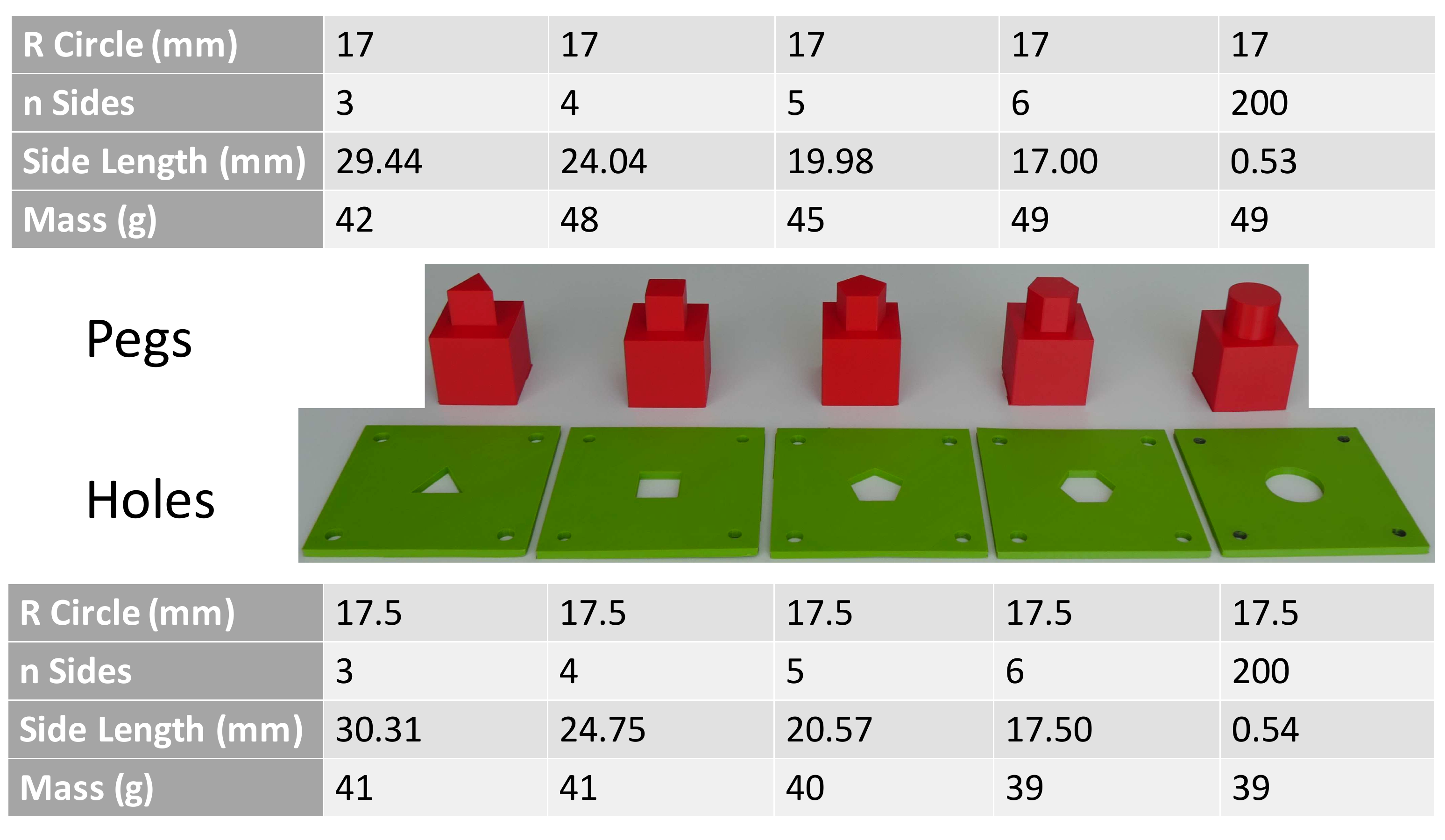}
\caption{Pegs and holes used to acquire the data}
\label{fig:dataset_multishape}
\end{figure}

The dataset records object positions and interaction forces for a set of polyhedral pegs in contact with holes (see Fig.~\ref{fig:dataset_multishape}).
The face of the polyhedron in contact with the environment is n-gon regular convex polygons with $n=\{3,4,5,6,200\}$ in Fig.~\ref{fig:dataset_multishape}.

\begin{figure}[t]
\centering
\includegraphics[width=1.0\columnwidth]{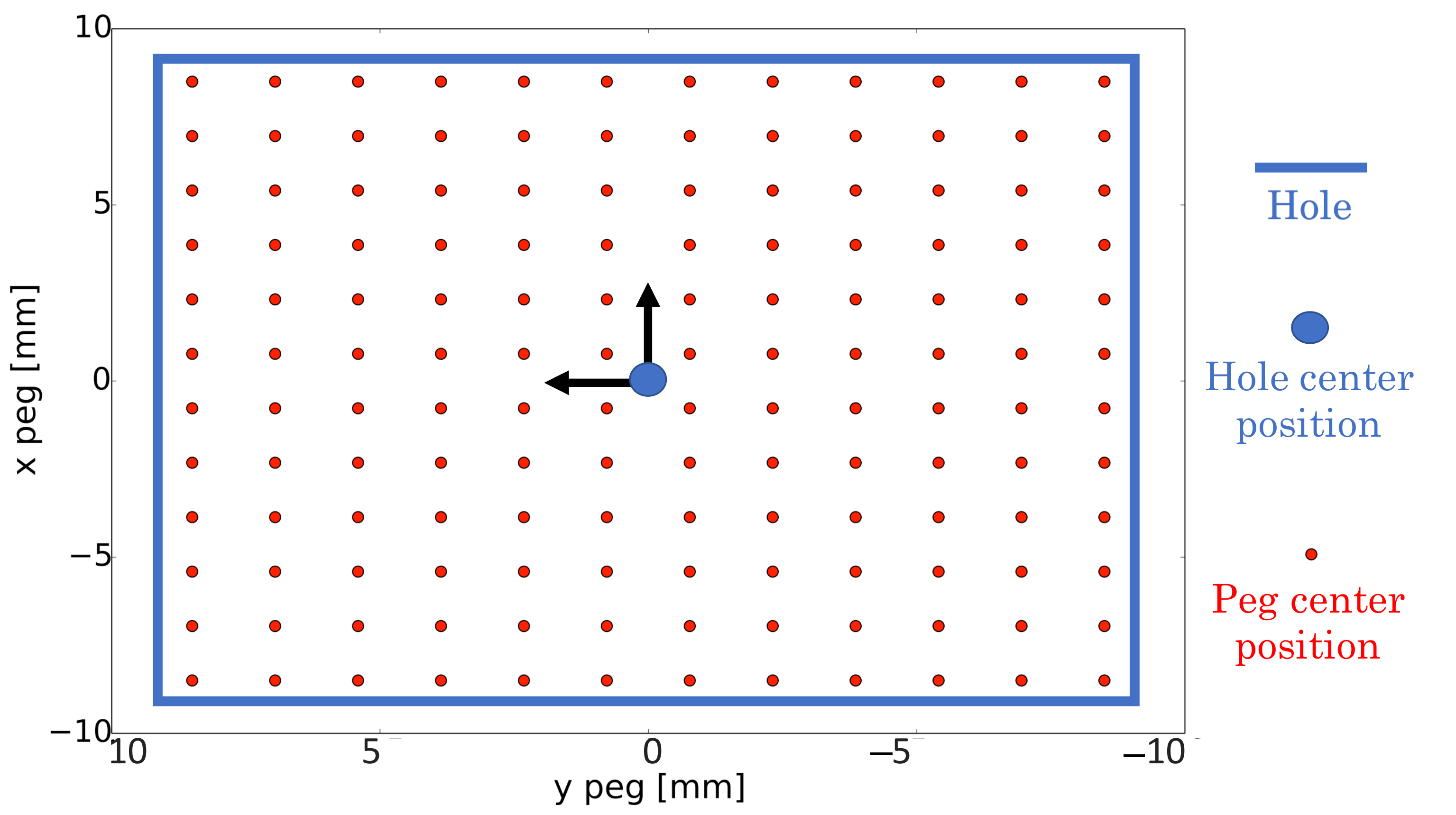}
\caption{Datapoints.}
\label{fig:dataset_points}
\end{figure}

The data was collected by sending a sequence of robot commands:
i) pick the peg, ii) rotate it in the given direction, iii) move to the center of hole with a predefined offset in $x$ and $y$ directions, iv) pushes the peg against the hole plate with a downwards force of \SI{30}{\newton} for \SI{10}{\second}.
Pushing the plate for a long time help the controller passing form the transient to the steady situation.
The force control is executed and recorded at \SI{100}{\hertz}.
The value of the force torque sensor and the end effector position for each point is recorded and stored as vectors of $\{Fx,Fy,Fz,Mx,My,Mz,Px,Py,Pz,Az,t,cont\}$.
$Px$, $Py$, $Pz$, are the peg positions with respect to the hole; $Az$ is the peg angle with the respect to the hole angle; $Fx$, $Fy$, $Fz$, $Mx$, $My$, $Mz$ are the forces and moments in the force sensor frame; $t$ is the time and $cont$ is a counter of the datapoints in Fig.~\ref{fig:dataset_points}.

\begin{figure}[b]
\centering
\includegraphics[width=1.0\columnwidth]{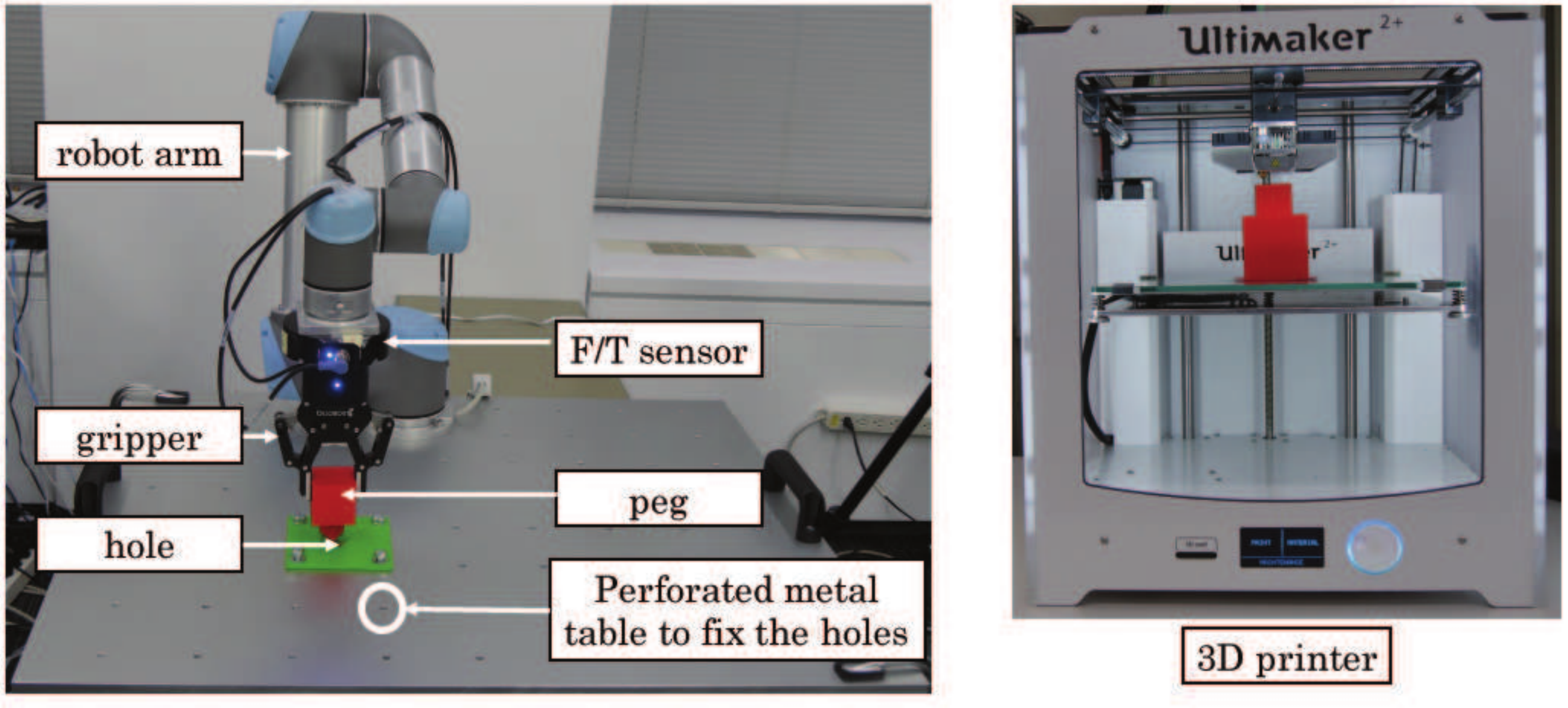}
\caption{Setup of the experiments.}
\label{fig:dataset_hw}
\end{figure}

As we are using a precisely calibrated table with a grid of screw holes, we can know the exact position of the hole.
To ensure that the relative position of peg and the hole are known accurately, we start each data collection by manually inserting the peg in the hole at correct orientation, making both position and angle of the peg aligned with the hole.
In this way, grasp errors do not have any effect on the experiments.
The position of the force sensor with respect to the peg position is shown in Fig.~\ref{fig:dataset_hw}.
The peg is moved by increments of \SI{1.5}{\milli\metre} in $x$ and $y$ direction within the range of $\pm$\SI{8.5}{\milli\metre} from the center of the hole (see Fig.~\ref{fig:dataset_points}).

Along with the dataset we also release the 3D models of all the objects used for collecting data. Details about the structure of the dataset are given inside the folders.

The dataset is available at \\
\url{http://ibm.biz/multishapeinsertion}

\subsection{Hardware}
Fig.~\ref{fig:dataset_hw} shows the setup for collecting the data. 
\begin{itemize}
\item \textit{Robot}: The system uses a UR5 industrial robotic arm with 6 DOF to precisely control the position of its tool center point (TCP). The robot has accuracy of $\pm$\SI{0.1}{\milli\metre}.
\item \textit{Gripper}: a Robotiq 2-finger 85 gripper is used to collect the dataset.
\item \textit{F/T sensor}: Robotiq force torque sensor FT-150 with effective resolution of \SI{0.2}{\newton} for the force and \SI{0.02}{\newton\metre} for the moment. Signal noise of \SI{0.5}{\newton} for the force and \SI{0.03}{\newton\metre} for the moment. To remove long-term drift, we recalibrate the force sensor.
\item \textit{Objects}: The objects (pegs and holes) are printed using an Ultimaker 2+ 3D printer using PLA \SI{0.75}{\milli\metre} filament with nozzle size \SI{0.4}{\milli\metre} and infill density of 20\%.
\end{itemize}

\subsection{Software}
The UR5 robot has many components available in the Robot Operating System (ROS) framework.
We use such ROS nodes to collect our data.
The F/T captured data are published as ROS topics and recorded at 100 Hz.
The object position with respect to the hole is only given before the peg goes in contact with the hole.

As we are in contact with the environment during the search phase and alignment, we adopt a common admittance control to stabilize the interaction betweeen robot and environment.
This controller is common for many industrial manipulators controlled by a position controller~\cite{article:robotica:freund98}.



\section{A Method for Labeling the Data for Multi-Shape Insertions}
\label{sec:sup}

Here we illustrate a method to label each entry of the dataset for performing the peg-in-hole task.
In the next section, we show how we can label each entry of the dataset for the shape recognition task.

During the execution of the peg-in-hole task, the position and orientation of the hole are inaccurate due to different uncertainties. Using the accurate position and orientation of our dataset, we can calculate for each entry the best action that should be performed. The action is the label and the input is the force and moments.
The position and orientation are then only used for labeling the data.

\begin{figure}[t]
\centering
\includegraphics[width=1.0\columnwidth]{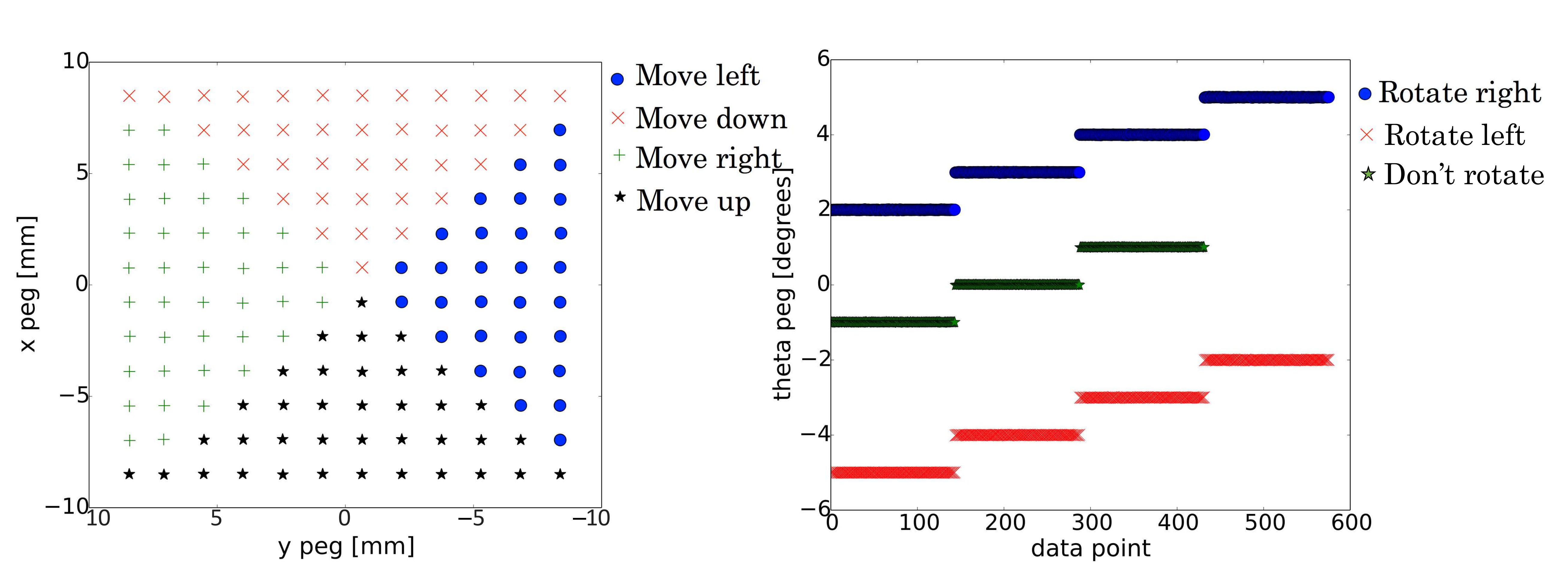}
\caption{Labels based on peg position and rotation for search phase.}
\label{fig:labels}
\end{figure}

Using only the peg positions $\left(x_{p},y_{p}\right)$, we label each entry of the dataset in 4 actions: move left, move down, move right and move up (see Fig.~\ref{fig:labels}). These actions allows to reduce the position error of the peg respect to the hole during the search phase.

%

To reduce the orientation error of the peg respect to the hole, we use the following labels:

\begin{tabular}{@{}r @{}c @{}c @{}c @{}r l}
$\theta_{p}$  & & $\ge$ & & $\theta_{l}$ & rotate right the peg \\
$\theta_{p}$ & & $\le$ & & $-\theta_{l}$ & rotate left the peg \\
$-\theta_{l}$ & $ > $ & $ \theta_{p} $ & $ < $ & $\theta_{l}$ & don't rotate \\
\end{tabular}

where $\theta_{p}$ is the peg orientation respect to the hole and $\theta_{l}$ is manually defined as a function of the clearance between the peg and hole.


\section{Analysis and Experiments}
\label{sec:results}

\subsection{Analysis of Force Data}
\label{sec:contact_situation}
\begin{figure}[t]
\centering
\includegraphics[width=1.0\columnwidth]{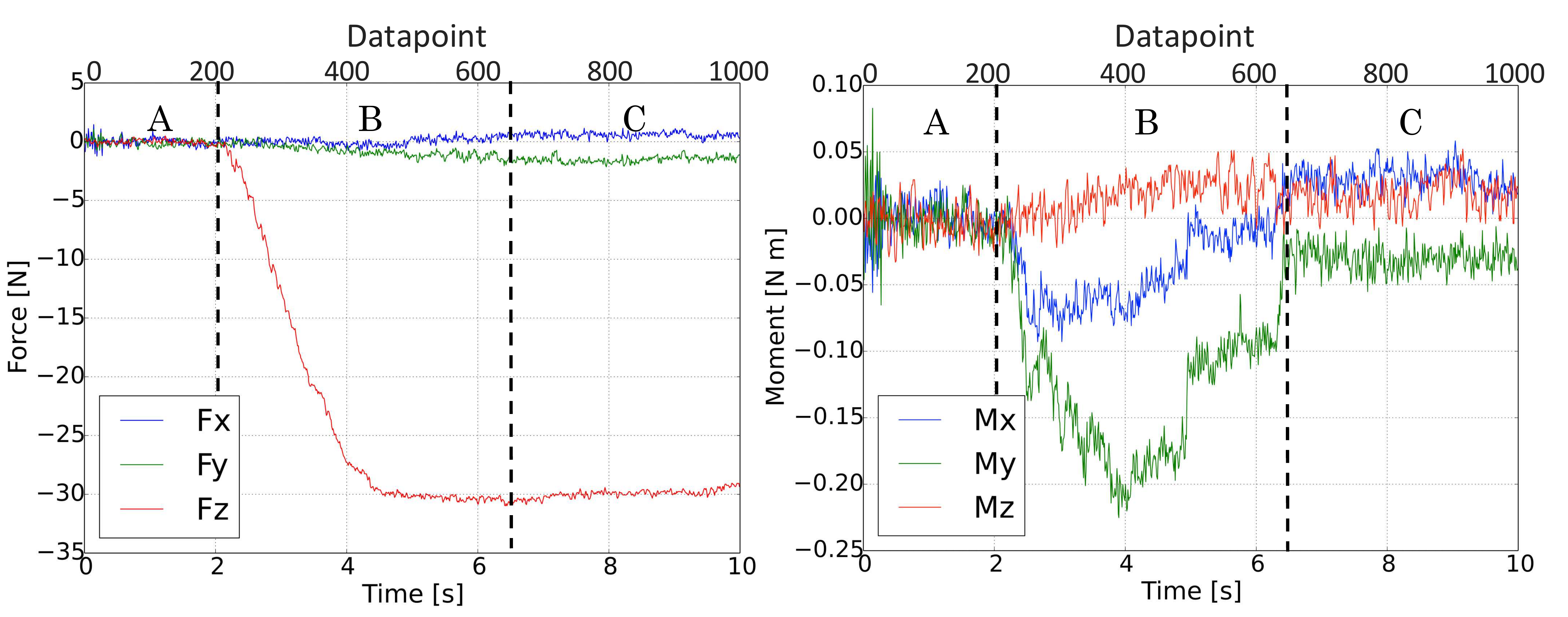}
\caption{Force and moment during the top left point in Fig.~\ref{fig:force_mom_data} using a peg with $n=4$}
\label{fig:force_mom_data}
\end{figure}

Fig.~\ref{fig:force_mom_data} shows the forces and moments of the dataset for a peg with $n=4$ at the top left position in Fig.~\ref{fig:dataset_points}.

In Fig.~\ref{fig:force_mom_data}, we can clearly distinguish three phases:

\begin{enumerate}[label=\Alph*.]
\item Non-contact: the peg is not in contact with the environment. This situation is not interesting for analyzing the contact and the insertion.
\item Transient: the forces and moments keep changing with time. In this period, we can analyze the response of the robot to the interaction with the environment.
\item Steady: the forces and moments of interactions remains almost unchanged in time.
\end{enumerate}

In the next sections, we will start to analyze situations B and C separately.
Lastly, we also analyze the combination of the data from both situations.

\subsection{Comparaison of Classifiers}


We train multiple classifiers using different methods with a cross entropy cost function. The classifier aims to find the best action for the given input (forces and moments).
Note that the accuracy of the classifier is the accuracy to provide a correct action given the forces and moments as inputs and is not the probability of entering the hole.



To compare the classifiers using different machine learning (ML) methods, we prepare the training data for each method by the following procedure. First, we sample raw data from a time window starting at the index $l$ and ending at the index $u$. We separate the data to frames each of which has the length of $d$. That results in $m = (u-l+1)/d$ frames. In each frame, we calculate the average of the raw data. We use the obtained $m$ average data as a training data.



We compare the accuracy of the ML methods of two tasks using the label explained in Sec.~\ref{sec:sup}; (T1) reduce the position error between peg and hole position using 4 force actions, and (T2) reduce the position error between peg and hole position using 3 moment actions.

In Table~\ref{tab:comparaison_method}, we compare the classifiers using only shape $n=4$ and the following 4 inputs:
\begin{equation}
\resizebox{0.91\hsize}{!}{%
$\text{input} = \left[\frac{\boldsymbol{F_x}}{\max(\left|\boldsymbol{F_x}\right|)},\frac{\boldsymbol{F_y}}{\max(\left|\boldsymbol{F_y}\right|)},\frac{\boldsymbol{M_x}}{\max(\left|\boldsymbol{M_x}\right|)},\frac{\boldsymbol{M_y}}{\max(\left|\boldsymbol{M_y}\right|)}\right]$
}
\label{eq:input}
\end{equation}
with $d=800$, $l=200$, $u=1000$. We made the comparison using the following techniques: SVM is support vector machine classifier with linear kernel, DT is the Decision Tree, RNDF is the Random Forest method, ADA is Ada Boost classifier, GAUS is the Gaussian Naive Bayes method, LDA is a Linear Discriminant Analysis, QDA is a Quadratic Discriminant Analysis and MLP is the Multi Layer Perceptron.

In Sec.~\ref{sec:diff_inputs}, we also compare the results of adding the remaining 2 F/T inputs.
 
\begin{table}[htpb]
\caption{Comparison of the different machine learning techniques for labels based on peg position with 4 actions (T1) and for labels based on peg orientation with 3 actions (T2).}
\label{tab:comparaison_method}
\begin{center}
\begin{tabular}{|c|c|c|c|}
\hline
\textbf{Technique} &\textbf{Acc [\%]  - T1} &\textbf{Acc [\%]  - T2} & \textbf{Average}\\ 
\hline
SVM & $62.66$ & $33.54$ & $48.1$\\
\hline
DT & $63.92$ & $46.52$ & $55.22$\\
\hline
RNDF & $68.67$ & $45.57$ & $57.12$\\
\hline
ADA & $64.87$ & $44.62$ & $54.74$\\
\hline
GAUS & $60.76$ & $40.82$ & $50.79$\\
\hline
LDA & $63.92$ & $40.19$ & $52.05$\\
\hline
QDA & $62.97$ & $37.97$ & $50.47$\\
\hline
MLP & $68.67$ & $53.16$ & $60.91$\\
\hline
\end{tabular}
\end{center}
\end{table}
From Table~\ref{tab:comparaison_method}, MLP is the best choice in both tests. The MLP network is composed of 2 hidden layers of size [100, 50], the optimizer is lbfgs and activation function is rectified linear unit.
In the next sections, we will only use this MLP network.

\subsection{Study of the different contact situations}
In Table~\ref{tab:comparaison_contact}, we compare the accuracy for the transient and steady state.

\begin{table}[htpb]
\caption{Comparison of the different contact situations using MLP}
\label{tab:comparaison_contact}
\begin{center}
\resizebox{\columnwidth}{!}{%
\begin{tabular}{|c|c|c|c|c|c|c|}
\cline{2-7}
 \multicolumn{1}{r|}{}  &$d$ &$l$ &$u$ &\textbf{Acc [\%] - T1} &\textbf{Acc [\%] - T2} &\textbf{Situation}\\  
\hline
a) &400 &200 &600 &69.62 &56.33 &Transient\\ 
\hline
b) &400 &600 &1000 &67.09 &46.52 &Steady\\ 
\hline
c) &800 &200 &1000 &68.67 &53.16 &Both\\
\hline
d) &50 &200 &600 &70.57 &61.39 &Transient\\
\hline
e) &50 &600 &1000 &65.51 &45.25 &Steady\\ 
\hline
f) &50 &200 &1000 &72.47 &54.75 &Both\\ 
\hline
\end{tabular}
}
\end{center}
\end{table}

In Table~\ref{tab:comparaison_contact}, we noticed that, for the task T2, taking only the average of the data points during the transient situation, the accuracy improved from c) $53.16\%$ to a) $56.33\%$.
On the other hand, by taking the average of only steady situation the results decrease to b) $46.52\%$.
From this result, we can suppose that the information about the task T2 is mostly during transient situation.

Another important result for T2 coming from the analysis of the dataset during the transient situation is that using the following parameters ($d=50$, $l=200$, $u=600$), the accuracy increases to d) $61.39\%$. The input is a sequence of 8 data points, i.e. [(600-200)/50=8].

For T1, the accuracy considering only the transient situation increase to a) $69.62\%$ and using only the steady situation the accuracy decreases to b) $67.09\%$. Using a sequence of 8 points as input during the transient situation the accuracy increase to d) $70.57\%$. As shown in Table~\ref{tab:comparaison_contact} for ($d=50$, $l=200$, $u=1000$), the steady contact situation is the most important phase for T1 and the accuracy increases to f) $72.47\%$).

During the steady situation using the following parameters ($d=50$, $l=600$, $u=1000$), the accuracy decrease to e) $65.51\%$ for T1 and to e) $45.25\%$ for T2.

Analyzing these results, we can affirm that the dynamic during the impact between the peg and environment is very important to understand the insertion task for the search and alignment phases.

We can conclude that for our analysis using MLP the main information for T2 is the transient, while for T1 is the whole contact phase. The parameters ($d=50$, $l=200$, $u=1000$) in Table~\ref{tab:comparaison_contact} are a good compromise and we choose these parameters to analyze the results for different shapes.

\subsection{Study of different inputs}
\label{sec:diff_inputs}
Another important analysis of the dataset is to understand which inputs are the most important. Adding $F_z$ as input the accuracy decrease from $70.57\%$ to $69.19\%$, adding $M_z$ the accuracy is $70.03\%$.
Therefore, the main information is in $F_x$, $F_y$, $M_x$, $M_y$.

\subsection{Study of different shapes}

Table~\ref{tab:comparaison_shapes} shows the results for the different shapes. From the table, we can clearly understand that T1 is easier than T2. In particular, we can notice that while the accuracy for the T1 increase with the number of sides, the accuracy for T2 is similar for all shapes. 

\begin{table}[htpb]
\caption{Comparison of the different shapes using ($d=50$, $l=200$, $u=1000$)}
\label{tab:comparaison_shapes}
\begin{center}
\begin{tabular}{|c|c|c|c|c|c|c|}
\hline
\textbf{n sides} &\textbf{Acc [\%] - T1} &\textbf{Acc [\%] - T2}\\  
\hline
3 			&67.09 			&67.72\\ 
\hline
4 			&72.47 			&54.75\\
\hline
5 			&72.78 			&65.19\\
\hline
6 			&75.00 		   &65.51\\ 
\hline
200 		&81.01 			&62.66\\
\hline
\end{tabular}
\end{center}
\end{table}


\subsection{Robot Experiments}
The model learned using the dataset is used to perform the task on UR5 robot.
As input we use the input of Eq.~\eqref{eq:input} with $d=50$, $l=200$, $u=1000$.
We use the following 4 actions during T1: $[F_x,0]$, $[0,F_y]$, $[-F_x,0]$, $[0,-F_y]$. And 3 actions during T2 $[M_z,]$, $[-M_z]$, $[0]$. For the experiment, we fix $F_x = 20$\SI{}{\newton}, $F_y = 20$\SI{}{\newton}, $M_z = 1$\SI{}{\newton\metre}.
Using this parameters the robot perform the insertion task in average after $7.5$ actions starting from a distance of the hole of \SI{2}{\milli\metre} with 100\% success rate. These results depend on the amplitude of the force and moment commands.
 
The video is available at \\
\url{https://youtu.be/6rLc9fAtzAQ}

In the video, the robot used the learned model to perform the insertion for all shapes.

\subsection{Shape Recognition}
We use the force and moment during contact to recognize the shape of the peg and the hole.
In our dataset, peg and the hole have the same shape.
We label the data using 5 classes (one per shape).
Using MLP and $d=50$, $l=200$, $u=1000$, we obtain an accuracy of $85.34\%$.

The result shows that using our dataset, the robot can also recognize the shape of the peg.
If the robot has low confidence that it is holding the correct peg, it can generate an error with the reason of failure.

\section{Conclusions}
\label{sec:conclusions}
In this paper, we presented a dataset for multishape peg-in-hole. Using this dataset, we conducted several analysis and we trained a MLP network able to select the right action based on forces and moments. The learned motion was tested on the UR5 robot. 


In a near future, we would like to work with deeper hole. Moreover, the current data set does not consider angular alignment errors except for the rotation about the peg axis. We will investigate more in this direction.
Another interesting area for future works would be transfer learning where the models are learned in simulation and fine-tuned on the real robot or where the insertion is learned from plastic pegs-holes and used with metal objects.


\small
\bibliographystyle{IEEEtran}
\bibliography{2018-rsj.bib}
\normalsize
\end{document}